\documentclass[runningheads]{llncs}
\usepackage[utf8]{inputenc}
\usepackage{graphicx}
\usepackage{tabularx}
\usepackage{array}
\usepackage{multirow}
\usepackage{arydshln}
\usepackage{dblfloatfix}
\usepackage{tikz}
\usepackage{calc}
\usepackage{amsmath}
\usepackage{mathtools}
\usepackage{amsfonts}

\usepackage{listings}
\usepackage{xcolor}
\usepackage{caption}
\usepackage{subcaption}
\DeclareCaptionFont{white}{\color{white}}
\DeclareCaptionFormat{listing}{\colorbox[cmyk]{0.43, 0.35, 0.35,0.01}{\parbox{\textwidth}{\hspace{14pt}#1#2#3}}}
\begin{document}
\title{OAK4XAI: Model towards Out-Of-Box eXplainable Artificial Intelligence \\for Digital Agriculture}

%\subtitle{for Dynamic Knowledge Representation and Management}
\author{Quoc Hung Ngo \and Tahar Kechadi \and Nhien-An Le-Khac}
%

\institute{School of Computer Science, College of Science\\
University College Dublin, Belfield, Dublin 4, Ireland\\
\email{hung.ngo@ucdconnect.ie},\\
\email{tahar.kechadi@@ucd.ie},\\
\email{an.lekhac@ucd.ie}}
\maketitle             % typeset the header of the contribution
\begin{abstract}
Recent  machine   learning  approaches   have  been  effective   in  Artificial   Intelligence  (AI)
applications. They  produce robust results  with a  high level of  accuracy. However, most  of these
techniques  do  not provide  human-understandable  explanations  for  supporting their  results  and
decisions. They usually act as black boxes, and it is not easy to understand how decisions have been
made. Explainable Artificial Intelligence (XAI), which has received much interest recently, tries to
provide human-understandable explanations  for decision-making and trained AI  models. For instance,
in digital  agriculture, related domains often  present peculiar or  input features with no  link to
background knowledge.  The application  of the  data mining  process on  agricultural data  leads to
results (knowledge), which are difficult to explain.

In this paper, we propose a knowledge map model and an ontology design as an XAI framework (OAK4XAI)
to deal with this issue. The framework does not only consider the data analysis part of the process,
but  it takes  into account  the semantics  aspect of  the domain  knowledge via  an ontology  and a
knowledge map model, provided  as modules of the framework. Many ongoing XAI  studies aim to provide
accurate and verbalizable accounts  for how given feature values contribute  to model decisions. The
proposed approach, however, focuses on providing consistent information and definitions of concepts,
algorithms,  and values  involved in  the data  mining models.   We built  an Agriculture  Computing
Ontology (AgriComO) to explain the knowledge mined in agriculture.  AgriComO has a well-designed structure and
includes  a wide  range  of concepts  and  transformations suitable  for  agriculture and  computing
domains.

. 
\end{abstract}

%-------------------------------------------------------------------------------------------------
\section{Introduction}
\label{sec:Intro}

Artificial Intelligence (AI)  applications are present in many domains  nowadays. These applications
have a  direct impact on human  lives, such as  healthcare, self-driving vehicles, smart  homes, the
military, etc. The advances in AI and big data have led to the rise of explainable AI (XAI) and have
gained much attention in recent years. Several studies have provided the main concepts, motivations,
and      implications      of      enabling     explainability      in      intelligent      systems
\cite{adadi2018peeking,arrieta2020explainable,gunning2017explainable}.  Other  studies have provided
an overview  of the existing XAI  approaches and future  XAI research opportunities. The  concept of
explainability is  closely related to interpretability.   XAI systems are interpretable  if they can
make human-understandable operations and decisions.
Some previous  studies considered the  XAI goals as  the need for  interpretable AI models,  such as
trustworthiness, causality,  transferability, informativeness, fairness,  accessibility, confidence,
interactivity,                      and                       privacy                      awareness
\cite{adadi2018peeking,arrieta2020explainable,gunning2017explainable}.  Some others  synthesized the
definitions  for the  XAI  goals and  provided  a  set of  WH-questions  to classify  explainability
approaches,       including      what,       who,       why,      what       for,      and       how
\cite{akula2019natural,arrieta2020explainable}.

XAI approaches can be  used to explain one of the three stages:  pre-model, in-model, and post-model
\cite{adadi2018peeking,molnar2020interpretable}. Pre-modelling targets a better understanding of the
datasets, while  post-model aims at  model approximation and  reporting of the  results. Explainable
modelling focuses on understanding how an AI model makes decisions. Many recent studies have focused
on explainable modelling; however, it is not  enough to understand the decision-making process in AI
applications. 

In  agriculture, AI  applications  are  constantly growing  at  a very  high  rate  during the  last
decade. These include soil studies, weather forecasting, crop yield prediction, disease predictions,
etc. For instance, there are several soil studies, building soil profiles \cite{shangguan2013china},
monitoring  soil   characteristics  under  the  effects   of  other  factors  and   crop  yield\cite
{bishop2001comparison},  or  using  soil  characteristics  to  predict  other  soil  characteristics
\cite{wang2019comparison}. 
Although  the number  of  XAI studies  in  digital agriculture  is  not too  high,  XAI for  digital
agriculture is necessary because agronomists and farmers  do not have a strong background in machine
learning and AI.  \cite{tsakiridis2020versatile} proposed an explainable AI  decision support system
based on  fuzzy rules to automate  field irrigation. Moreover,  even though the number  of knowledge
models is small, the input and output attributes of forecasting models are different, and the number
of agricultural  features is large  and diversified. The majority  of these results  (knowledge) are
stored  as pre-trained  models,  computer software,  or scientific  papers/reports.  They also  lack
explanations to assist different users in accessing and understanding them. 

In  this paper,  we  propose an  ontology-based  knowledge map  \cite{ADMIRE,ngo2022knowledge}  for representing  and
explaining the  mined knowledge,  which has  been produced  previously by  the data  mining process,
including classification, regression, clustering, association rules,  and other forms of mining. The
main contribution of the model as an out-of-box approach for DM models is to support data scientists
and agronomists in managing, understanding, and using mined results/knowledge for decision-making.

The next  section overviews the  OAK model as  a foundational theory  for the proposed  approach and
details the core of the ontology used in this study. Section \ref{sec:OAK4XAI} presents the proposed
ontology-based  knowledge  map  model  for  explainable  AI,  including  its  architecture  and  XAI
transparency.We describe the  implementation process for validating the proposed  model based on the
knowledge repository  in digital agriculture  in Section \ref{sec:Validation}. Finally,  we conclude
the paper and give some future research directions in Section \ref{sec:conc}.

%-------------------------------------------------------------------------------------------------
\section{OAK - Ontology-based Knowledge Map Model}
\label{sec:OAK}

\subsection{Knowledge Definitions}

We introduced the ontology-based Knowledge Map Model  (OAK) to handle the knowledge extracted from a
DM process \cite{ngo2020oak,ngo2022knowledge}. Before  going into the details of the proposed  approach, we give some
key  definitions of  the OAK  model.  The model  includes {\it  knowledge representation,  ontology,
  knowledge map, concept, attribute, transformation, instance, state}, and {\it relation}. 

Knowledge represents the result of an experience or  a data mining process, which uses some learning
algorithms to  predict a target  based on the  input dataset. Knowledge  is of two  types: processed
knowledge and  factual knowledge.  The {\it  Processed Knowledge}  represents the  result of  a data
mining process. This knowledge has some attributes that characterise it; input and output attributes
and  learning  algorithms. {\it  Factual  Knowledge}  is information  validated  by  experts in  the
domain. It is characterised  by some attributes, such as the transformation of  input to output, its
states/values, etc.

In  the OAK  model,  the processed  knowledge  can  be of  four  types: classification,  regression,
clustering,  and association  rule \cite{ngo2020oak,ngo2022knowledge}.   {\it Regression}  is a  ML
function  that  predicts  a continuous  outcome  variable  based  on  the values  of  the  condition
variables. This means that the {\it  Regression} model (\(k_{Reg}\)) uses regression type algorithms
(\(t\), \(t \in \mathbb{T}_{DM}\), and \(t\) to predict  the value of the attribute ({\it target} of
the    model)   based    on   its    input    attributes    ({\it   conditions}   of   the    model)
\cite{ngo2022knowledge}. These are defined as follows: 

\begin{equation*}
    \begin{small}
    \begin{aligned}
     k_{Reg} = & (\{i\}, \{r\}, \{t\}, \{s\}) \\
     \{i\} = & i_{regressor} \cup \{i_{condition}\} \cup \{i_{target}\} \cup \{i_{dataset}\} \cup \{i_{evaluation}\} \\
     \{r\} = & \{ (i_{regression}, hasAlgorithm, t_{DM})\}\; \cup \\
     & \{ (i_{regressor}, hasRegressor, i_{regressor})\}\; \cup \\
     & \{ (i_{regressor}, hasCondition, i_{condition})\}\; \cup \\
     & \{ (i_{regressor}, predicts, i_{target})\}\; \cup \\
     & \{ (i_{condition}, hasTransformation, t_{D})\}\; \cup \\
     & \{ (i_{target}, hasTransformation, t_{D})\}\; \cup \\
     \{t\} = & \{ t_{DM}=algorithm(c) \in \mathbb{T}_{DM}; c=Regression) \}\; \cup \\
     & \{ t_D=f(i): \Re_x \rightarrow \Re_y\ ; 
        \ i \in \{i_{condition}\} \} \; \cup \\
    \end{aligned}
    \end{small}
\end{equation*}
\begin{equation*}
    \begin{small}
    \begin{aligned}
     & \{ t_D=f(i): \Re_x \rightarrow \Re_y\ ; \ i \in \{i_{target}\} \} \\
     \{s\} = & \{ \forall s \in \mathbb{S}, \exists i \in \{i_{condition}\}: i 
    	\overset{hasState}{\rightarrow} s \}\; \cup \\
     & \{ \forall s \in \mathbb{S}, \exists i \in \{i_{target}\}: i 
    	\overset{hasState}{\rightarrow} s \} \\
    \end{aligned}
    \end{small}
\end{equation*}
The regressor  \(k_{Reg}\) is characterised by  its main components, which  are datasets, prediction
targets, conditions, and evaluation information. In addition, the processed knowledge can have other
attributes related to locations, research context, etc. 

In  summary, the  proposed model  includes all  the information  necessary to  characterise a  given
knowledge  (result) at  any  time within  a  data mining  application. The  model  has an  efficient
knowledge  representation   that  facilitates   its  storage  and   retrieval  from   the  knowledge
repository. However, it  still needs an explainable  mechanism to interpret the concepts  as well as
its processing steps. Therefore, an explainable knowledge base or a suitable ontology can be used in
interpreting the mined knowledge within a given DM application. 

\subsection{Ontology: Role and Design}
\label{sec:Ontology}

While the OAK model is designed to handle any  domain knowledge, its ontology is specific to a given
domain for  efficient exploitation.  One of  the main  objectives of  ontology is  to assist  in the
explanation and interpretation of  the mined results. In the proposed  approach, the ontology covers
three  main functions:  1) agriculture  common concepts  definition and  representation, 2)  concept
transformations handling, and 3) Representation of the main types of relationships between concepts.

\begin{figure}[h]
  \centering
  \includegraphics[width=8.5cm]{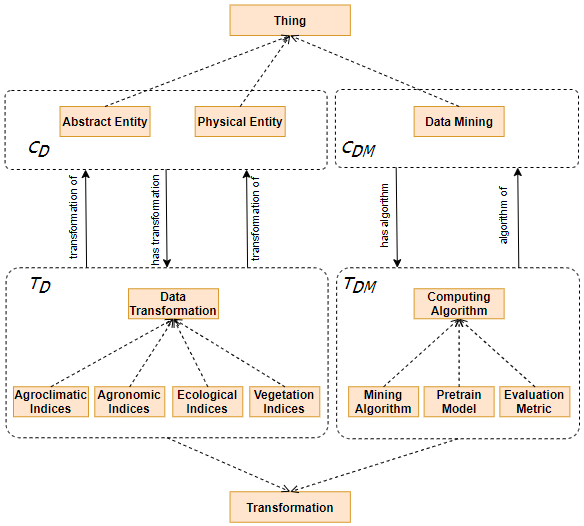}
  \caption{An overview of Agriculture Computing Ontology}
  \label{figAgriOntArchitecture}
\end{figure}

In this study,  we developed and implemented  an Agriculture Computing Ontology  (AgriComO) that contains
the most common classes (concepts), instances, attributes,  and relations in crop farming. The AgriComO’s
architecture is derived from the knowledge map (KMap) model and contains the following components: 

\begin{itemize} 
  \item  {\bf  Concepts:} Concepts  in  the  agriculture  {\it  field, farmer,  crop,  organization,
        location},  and {\it  product}.  The  DM concepts  include {\it  clustering, classification,
        regression}, and {\it association rule}.  
  \item  {\bf  Transformations:} are  predefined  transformation  functions  of agriculture  and  DM
        (See Figure \ref{figAgriOntArchitecture}).
  \item  {\bf Relations:}  They represent  relationships between  concepts/instances, and  they also
    represent the analysis process to create the knowledge. 
\end{itemize} 

\vspace{-2mm}

The AgriComO ontology describes agricultural concepts, their relationships, and lifecycles between seeds,
plants, harvesting, transportation, and consumption. The concept relationships concern weather, soil
conditions,  fertilizers,  and farms  description.  Moreover,  AgriComO  includes  DM concepts,  such  as
classification, clustering, regression,  and association rules. The combination  of agricultural and
DM concepts represents mined knowledge efficiently. For instance, in the current implementation, AgriComO
has    450    classes   and    over    3,381    axioms    related    to   agriculture    based    on
\cite{ngo2018ontology}. Finally,  it provides an overview  of the agricultural domain  with its most
general concepts. 

AgriComO is the core  ontology for building knowledge maps (KMaps) for digital  agriculture and adopts an
XAI-oriented  design at  the levels  of the  concepts and  relationships. Therefore,  every concept,
transformation, and relation  in the ontology has at  least two attributes; one for  the title ({\it
  rdfs:label}) and the  other for the description ({\it rdfs:comment}).  Moreover, we provided extra
attributes; {\it rdfs:isDefinedBy} and {\it rdfs:seeAlso} to provide external references for further
information for each concept.  These attributes are considered basic information  for each entity in
the  ontology.  The  definitions,  descriptions,  and comments  on  concepts,  transformations,  and
relations in the ontology provide transparency and explainability for AgriComO and the overall model. 

\begin{itemize}
  \item {\it URI} - Universal Resource Identifier.
  \item {\it rdfs:label} - name of concepts or instances.
  \item {\it rdfs:comment}  - description of concepts, instances, relations,  or transformations for
        explanation purposes. 
  \item {\it dc:identifier} - formula, expression, or function to calculate and transform data if it
        has. 
 \item {\it rdfs:isDefinedBy} - sources or creators of the concepts.
 \item {\it rdfs:seeAlso} - external references for further information for each concept.
\end{itemize}

In  summary, a  well-defined ontology  design  provides explainability  to  each concept  in the  DM
applications and each instance value in the knowledge representation and its processing. It supports
the OAK4XAI model (described  in the next section) in interpreting the  accurate meaning of concepts
and values in knowledge items and in helping users understand their decision-making. 

%----------------------------------------------------------------------------------------------
\section{OAK4XAI Model \& Architecture}
\label{sec:OAK4XAI}

\subsection{OAK4XAI Architecture}
\begin{figure}[ht]
  \centering
  \includegraphics[width=9cm]{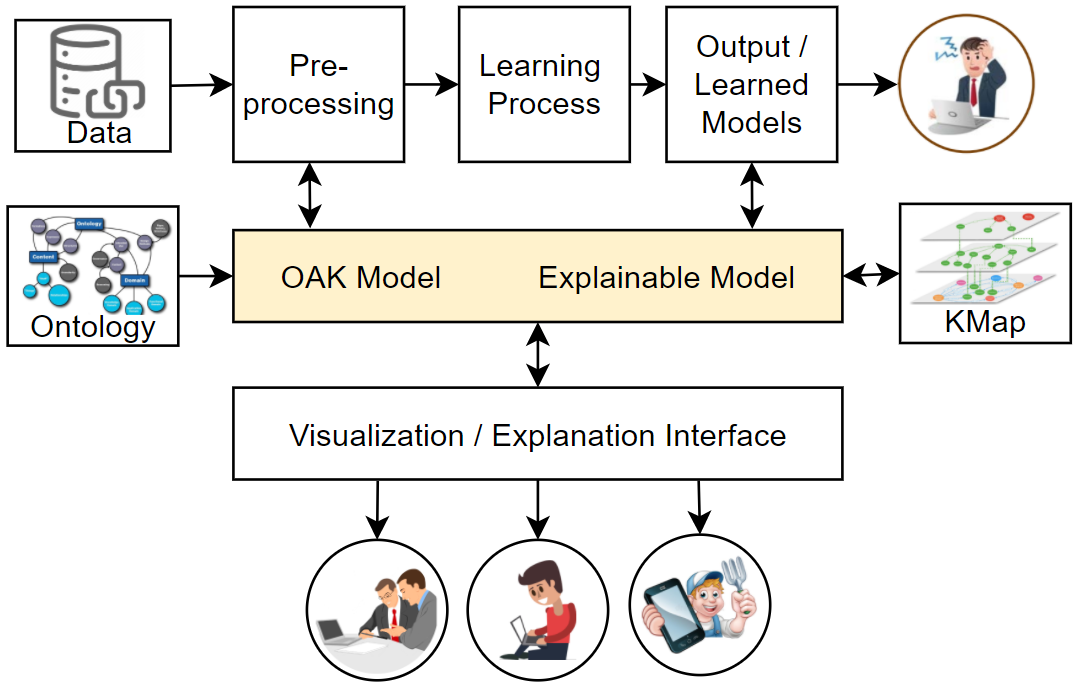}
  \caption{Architecture of OAK4XAI}
  \label{figOAK4XAImodel}
\end{figure}

The overall OAK4XAI  architecture is developed around OAK \cite{ngo2020oak}.  It organises knowledge
into  two  separate classes:  knowledge  and  its explanation.  The  knowledge  class manages  a  DM
application result (item) based on its relationships with other concepts and entities and is defined
in the  KMap module. The  knowledge explanations  are stored in  a pre-defined ontology  (See Figure
\ref{figOAK4XAImodel}).   This    architecture   handles   both   factual    and   mined   knowledge
\cite{ngo2020oak}. We use  a multi-layer approach where  knowledge items are in the  KMap, and their
explanations are located in the ontology. 

\setlength{\belowcaptionskip}{-10pt}

As  mentioned  in  Section  \ref{sec:Ontology},  AgriComO   includes  agriculture  concepts  and  the  DM
domain. Each concept contains necessary descriptions and  attributes to identify the concept and its
data processing methods. They are defined under  the prefix for AgriComO\footnote{URI prefix for concepts
  of {\it  AgriComO}: {\it http://www.ucd.ie/consus/AgriComO\#}} ontology.  All knowledge representations
(DM prediction models) in the agriculture  knowledge maps repository (AgriKMaps) are represented and
stored  as  KMaps  with a  prefix  for  the  AgriKMaps\footnote{URI  prefix for  instances  of  {\it
    AgriKMaps}: {\it http://www.ucd.ie/consus/AgriKMaps\#}} knowledge repository.
    
\subsection{Using OAK4XAI for Modeling and Explaining}
\label{sec:Modeling}

The proposed  approach converts  the mined  knowledge into  its corresponding  representation.  This
procedure  is  defined  in the  module  {\it  Knowledge  Wrapper}.   More precisely,  it  creates  a
representation for  the mined  knowledge using  the model $k  = (\{i\},  \{t\}, \{s\},  \{r\})$ (see
Section \ref{sec:OAK}) before converting it into  RDF turtles and importing them into the RDF Triple
storage.  This  consists of  six steps:  1) identify the  model; 2)  identify concepts;  3) generate
instances; 4) identify  transformations; 5) generate states; and 6)  generate scripts. The Knowledge
Wrapper implementation depends on the type of knowledge items, such as {\it Factual Knowledge} items
published  in scientific  papers  or {\it  Processed  Knowledge} items  extracted  directly from  DM
modules.

A representation $k$ includes a set of instances, transformations, states, and relations. The set of
instances $\{i\}$  is created  in Step 1  and Step 3,  while the  set of transformations  $\{t\}$ is
created in  Step 1 and Step  4. These transformations are  linked to instances to  represent the way
data is processed  in the prediction model. The set  of states $\{s\}$ is generated in  Step 5. Note
that not  all knowledge representations  have sets  of states.  If  the knowledge items  are factual
knowledge items, they  include values, and their representations contain  states. Otherwise, the set
of  states  is  empty.   Finally,  the  set   of  relations  $\{r\}$  is  based  on  {\it  rdf:type,
  AgriComO:hasTransformation,      AgriComO:hasState,      AgriComO:hasCondition},     and      {\it
  AgriComO:predicts}, etc.

\begin{lstlisting}[label=Regressor004,caption=Triples of Regressor\_004]
AgriKMaps:Regressor_004 
    rdf:type owl:NamedIndividual ,
             AgriComO:Regressor ,
             AgriComO:KnowledgeModel ;
    rdfs:label "Regressor 004" .
    AgriComO:definedIn 
            AgriKMaps:Article_004 ;
    AgriComO:hasAlgorithm 
            AgriComO:Algorithm_DTR ,
            AgriComO:Algorithm_LR  ,
            AgriComO:Algorithm_RF ,
            AgriComO:Algorithm_GBRT;
    AgriComO:hasCondition 
            AgriKMaps:SoilPH_004 ;
    AgriComO:predicts 
            AgriKMaps:SoilPH_004x ;
    AgriComO:hasDataset 
            AgriKMaps:Dataset_CONSUS_001 ;
AgriKMaps:SoilPH_004
    rdf:type owl:NamedIndividual ,
            AgriComO:SoilPH ,
    AgriComO:hasTransformation 
            AgriComO:Transformation_SoilPH_Max ,
            AgriComO:Transformation_SoilPH_Min ,
            AgriComO:Transformation_SoilPH_Avg ,
\end{lstlisting}

For  example,  when applying  this  modelling  process,  the  knowledge item  {\it  Regressor\_004},
published in  article {\it Article\_004}  \cite{ngo2019predicting}, can be  converted into a  set of
triples. The  brief triple  example of  {\it Regressor\_004}  (shown in  Listing \ref{Regressor004})
shows that  {\it Regressor\_004}  is used to  predict soil pH  and this  attribute uses the  soil pH
transformation    {\it   Transformation\_SoilPH\_Max,    Transformation\_SoilPH\_Min},   and    {\it
  Transformation\_SoilPH\_Avg}.  % 

The   semantic   interpretation  of   concepts,   transformations,   and   values  (such   as   {\it
  Transformation\_SoilPH\_Max})  in  this knowledge  item  is  done  through the  one-way  retrieval
process; from AgriKMaps to AgriComO and from AgriComO  to AgriComO. This means that using predefined information in
the AgriComO  ontology to interpret  instances and processing in  the AgriKMaps. The  explanation process
starts with  an instance in {\it  AgriKMap}. It browses the  knowledge item to identify  the concept
class (prefix {\it AgriComO}) and its attributes. 

\iffalse
\begin{lstlisting}[label=Explanation004,caption=Information for Soil pH attribute]
AgriKMaps:SoilPH_004x 
    rdf:type owl:NamedIndividual ,
            AgriComO:SoilPH ,
    AgriComO:hasTransformation 
            AgriComO:Transformation_SoilPH_Max ;
    AgriComO:hasTransformation 
            AgriComO:Transformation_SoilPH_Min ;
    AgriComO:hasState 
            AgriComO:SoilPH_StronglyAlkaline ;
Transformation_SoilPH_Tier11 
    rdf:type owl:NamedIndividual ,
            AgriComO:DataTranssformation ,
    rdf:comment "owl:NamedIndividual"^^xsd:string ,
\end{lstlisting}
\fi

\subsection{XAI Transparency with OAK4XAI}

\begin{figure*}[ht]
  \centering
  \includegraphics[width=\textwidth]{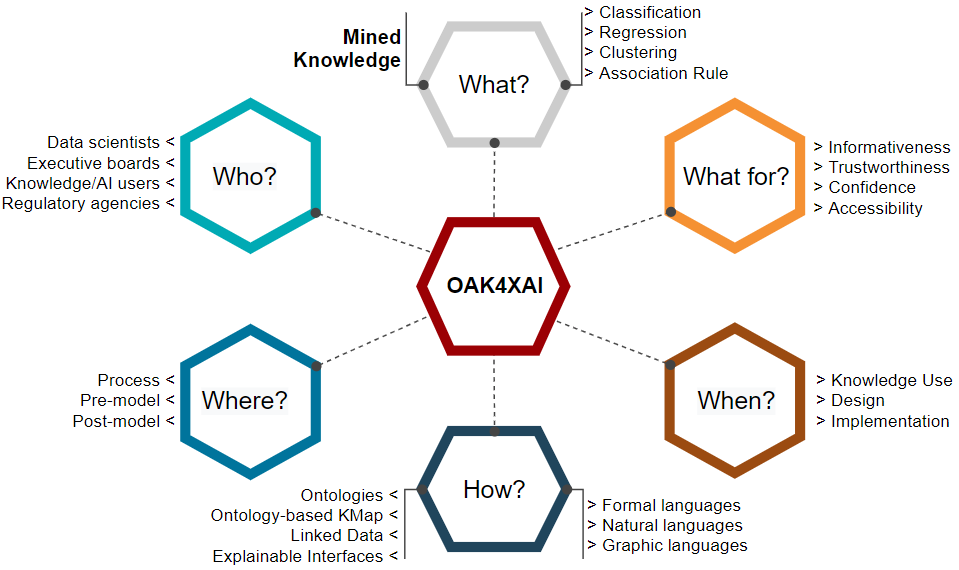}
  \caption{OAK4XAI Model for explainable AI}
  \label{figOAK4Explainability}
\end{figure*}

In this section,  we discuss the proposed model’s  XAI functions by answering a  set of WH-questions
(What,  What  for,   Who,  When,  Where  and  How)   along  with  a  summary  as   shown  in  Figure
\ref{figOAK4Explainability}. 

\setlength{\belowcaptionskip}{-10pt}

\paragraph{\bf What?} 
The OAK4XAI model focuses on representing and explaining  the results of the DM analysis process. It
supports four types of analyses: {\it  Classification, Clustering, Regression}, and {\it Association
  Rule}.

\paragraph{\bf What for?} 
OAK4XAI can assist in describing the meaning of  data, provide ways to transform raw data into input
data for learning models and explain the results of the learning models. Therefore, this model helps
to increase transparency, usability,  trust, and confidence in mined knowledge,  which are the XAI's
main goals. 

%\vspace{-5mm}
\paragraph{\bf Who?}
The  model  can explain  the  results  of  the ML  algorithms  to  different  types of  users.  More
precisely. The implemented version of this model supports the following user groups: 
\begin{itemize}
  \item {\it Managers and executive boards} understand an overview of knowledge items, including
        inputs, outputs, algorithms, training datasets, and results evaluation.
  \item {\it Data scientists and developers} understand attributes, values, and labels used in the
        knowledge items (DM models).
  \item {\it  Agronomists and experts}  understand states and  processing steps of  knowledge items,
        select then apply them in farming. 
\end{itemize}

%\vspace{-5mm}
%\vspace{-5mm}
\paragraph{\bf How?} 
Recall that  OAK4XAI contains a predefined  ontology, which helps  to explain the attributes  in the
knowledge representation. A representation  is an entry of KMap with  nodes for concepts, instances,
transformations, states,  and edges  for relationships  between them.  All  KMap entries  are stored
using the linked  data technique. Finally, explainable interfaces provide  different explanations to
different user groups. Moreover, the explanations need  to be written in suitable language (and with
a description strategy) for the intended audience. These include: 

\begin{itemize}
 \item {\it Formal language} can be structured or  syntax explanation, and it is based on structured
       and unique linked data in the knowledge repository layer. The linked data are sets of triples
       {\it (subject,  predictive, object)}. These triples  can be transformed and  interpreted in a
       certain format language. 
 \item {\it Natural  language:} written texts can be generated from the description  of concepts and
       instances  in  the  ontology.   Each  concept  in the  ontology  has  some  attributes  ({\it
         rdfs:label}  and  {\it  rdfs:comment}),  which   support  the  model  in  a  human-friendly
       explanation.  
 \item {\it Graphic  language:} visualizations can be supported by  the explanation interface layers
       based on the knowledge repository layer. 
\end{itemize}

%\vspace{-5mm}
\paragraph{\bf Where?} 
It is  worth noting  that OAK4XAI  can explain  data outside  the mining  steps, including  the {\it
  pre-model} and {\it post-model} stages of the DM applications. 
\begin{itemize}
  \item {\it Pre-model}: With this model, the users can pre-process and transform the data and input
        it into learning  models. Moreover, they have more options  for improving the pre-processing
        and  transformation steps  by using  predefined {\it  Transformation} instances  of the  AgriComO
        ontology. 
  \item {\it Post-model}: Trained  models and final decisions contain data  values, which need extra
        information and semantics.  At this stage,  different users may have different concerns (for
        example, why this decision was taken). This  can be solved by exploring detailed information
        about each concept and extra information.  
\end{itemize}

%\vspace{-5mm}
%\vspace{-5mm}
\paragraph{\bf When?} 
This model can be used to develop AI processes and XAI in three stages:
\begin{itemize}
  \item {\it  Designing:} planning pre-processing algorithms can  help understanding data  better by
        accessing predefined knowledge from the core ontology in the OAK4XAI model.
  \item {\it Implementation:} allowing scientists to have a deep understanding can contribute to the
        process of developing and improving machine learning models.
 \item  {\it Knowledge  Use:} explaining  knowledge results  can assist  in understanding  the model
        better. 
\end{itemize}

%-------------------------------------------------------------------------------------------------
\section{Validation}
\label{sec:Validation}

\subsection{Implementation}
OAK4XAI is implemented using the linked data technique, the graph database server, and the web-based
explanation application. The  graph database server supports RDF triple  storage and SPARQL protocol
for query,  while web-based explanation  application (including  {\it Knowledge Wrapper}  module and
{\it Knowledge  Browser} module for visualization).  The search engine searches  for knowledge items
from the  AgriKMaps domain,  while the explainable  engine retrieves the  domain knowledge  from the
predefined ontology, the  AgriComO domain. All instances and  concepts from AgriKMaps and AgriComO  
have a URI and they link together based  on   their   URI.   The   graph    database   uses   Apache
Jena\footnote{https://jena.apache.org/index.html}   (for  the   native  knowledge   graph  storage),
SPARQL\footnote{https://www.w3.org/TR/sparql11-query/}    1.1     (for    SPARQL     Engine),    and
Fuseki\footnote{https://jena.apache.org/documentation/fuseki2/}  (for  SPARQL Endpoint),  while  the
Knowledge Browser is a web-based application for exploring knowledge and providing explanations. 

\subsection{Explanation}
Knowledge Browser functions are twofold: it assists users in locating mined knowledge items based on
their  input queries  or  keywords and  validates  the  explainability of  the  DM analysis  process
results.

\begin{figure*}[h]
  \centering
  \includegraphics[width=\textwidth]{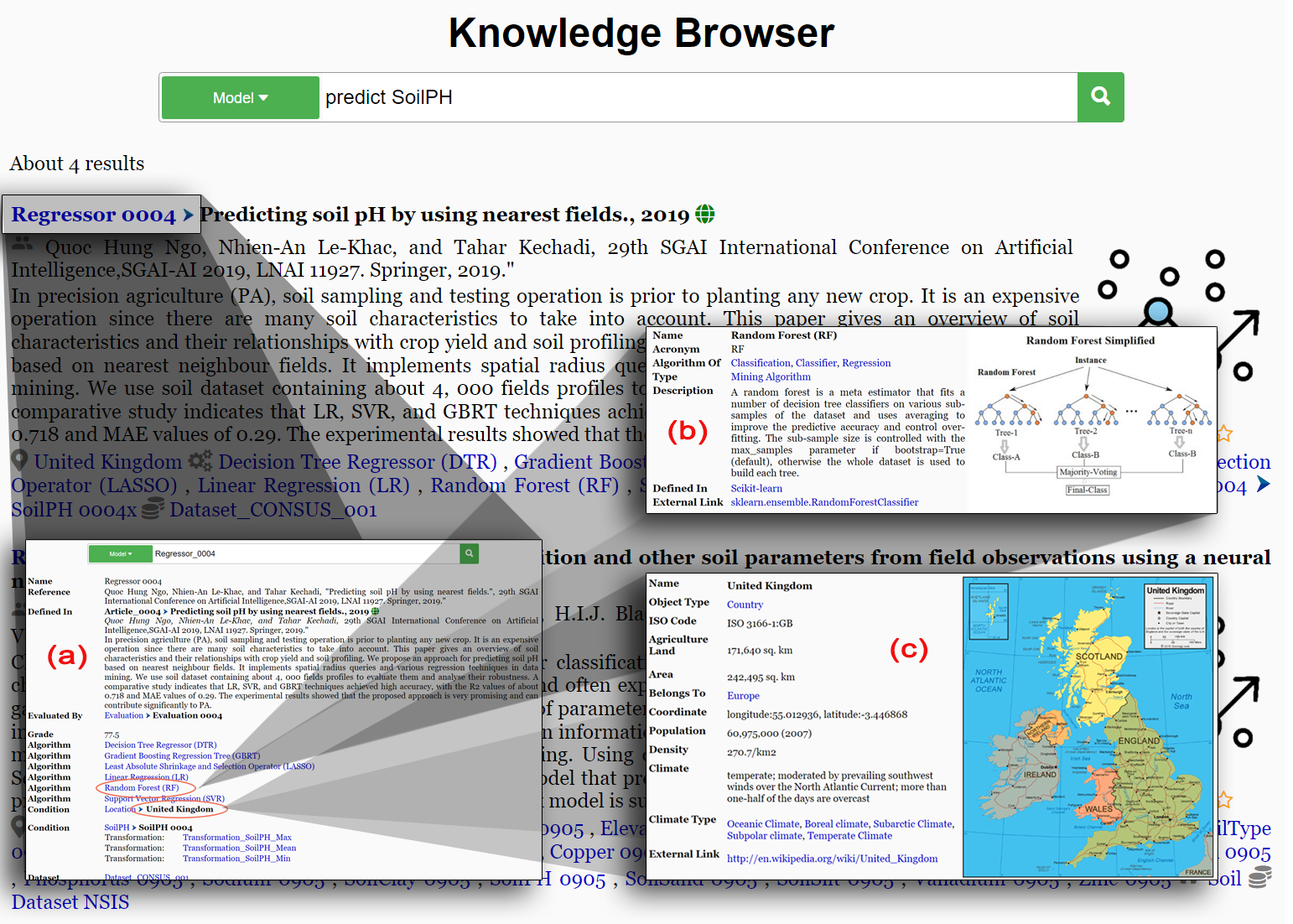}
  \caption{Explanation in Knowledge Browser}
  \label{figKnowledgeBrowserXAI}
\end{figure*}

Moreover, Knowledge Browser is a knowledge search engine, which looks for knowledge items (knowledge
maps) in  the RDF  storage and queried  input concepts  and their roles.  The retrieved  results are
represented and explained  in many levels of  detail, from general to details.  Knowledge Browser is
very efficient in finding  knowledge items in the knowledge repository based  on input queries, such
as "predict  soilPH" (with  the result  shown in  Figure \ref{figKnowledgeBrowserXAI}).  The process
includes: 

\begin{itemize}
  \item Finding knowledge items by search queries from {\it AgriKMaps};
  \item Segmenting concepts into parts AI models;
  \item Generating SPARQL queries;
  \item Generating summaries of knowledge items from return triples.
\end{itemize}

\setlength{\belowcaptionskip}{-15pt}

The explanation content  depends on users concerns when exploring  given items. The explanation
process with the OAK4XAI approach consists of the following steps:
\begin{itemize}
  \item Determine the concept in the knowledge item;
  \item Retrieve related attributes and descriptions from {\it AgriComO} ontology;
  \item Generate explanations interface (from return triples).
\end{itemize}

For example, to  locate the knowledge item  shown in Listing \ref{Regressor004}, the  results of the
query  "predict  SoilPH"  are  represented  as summaries  of  knowledge  items  (AgriKMaps)  (Figure
\ref{figKnowledgeBrowserXAI}a), and the  details of their concepts and states  are interpreted based
on knowledge from  the ontology (AgriComO) (Figure \ref{figKnowledgeBrowserXAI}b/c).  Details of the
data processing stage (formulas  and values) in the knowledge items are  described in the predefined
ontology, so their information can be represented  and explained. Based on different user questions,
the system has different levels of interpretations of the post-model stage of the knowledge items.  

Moreover, the system also supports searching for  concepts and related details in the ontology. This
explainability can assist users in the pre-model stage of building the knowledge items. For example,
Listing   \ref{TransformationSoilPHs}    provides   the    information   of    transformation   {\it
  Transformation\_SoilPH\_Max} using  in {\it Regressor0004} (Listing  \ref{Regressor004}) and other
transformations                        (such                         as,                        {\it
  Transformation\_SoilPH\_Tier11}\footnote{https://en.wikipedia.org/wiki/Soil\_pH}    (defined    by
US. Department of  Agriculture, Natural Resources Conservation  Service) as a category  of 11 types)
for the {\it soil pH} attribute. 

\clearpage

\begin{lstlisting}[label=TransformationSoilPHs,caption=Information of Transformations for Soil pH attribute]
[Transformation_SoilPH_Max] 
    Definition:
            This transformation returns the maximum value of  
            Soil pH values in the sampling area, for example in  
            the radius of 100m.
    Values:
            0.0 - 14.0 ;
[Transformation_SoilPH_Tier11] 
    Definition:
            This transformation returns a state of soil pH, 
            defined by the United States Department of 
            Agriculture Natural Resources Conservation Service, 
            classified soil pH ranges as 11 types.
    States:
            Ultra acidic	        < 3.5
            Extremely acidic	    3.5-4.4
            Very strongly acidic	4.5-5.0
            Strongly acidic	        5.1-5.5
            Moderately acidic	    5.6-6.0
            Slightly acidic	        6.1-6.5
            Neutral	                6.6-7.3
            Slightly alkaline	    7.4-7.8
            Moderately alkaline	    7.9-8.4
            Strongly alkaline	    8.5-9.0
            Very strongly alkaline	> 9.0
[...]
\end{lstlisting}

In summary, with a predefined AgriComO  ontology, OAK4XAI provides consistent information and definitions
of concepts,  algorithms, and values  involved in a  knowledge item. The  explainability information
enables  researchers to  communicate and  compare  the results  of various  classifiers and  support
stakeholders  in making  informed decisions  for the  implementation and  usage of  machine learning
models. 

\subsection{Statistics}
Numerous algorithms  and transformation  techniques have  been extracted  from several  resources to
support  the outside-of-box  explanation  of DM  applications in  agriculture.   They are  collected
manually from  programming libraries, scientific  articles, etc. A  list of algorithms  is collected
from   programming  libraries,   such  as   Scikit-learn\footnote{https://scikit-learn.org/stable/},
NLTK\footnote{https://www.nltk.org/},
Huggingface\footnote{https://huggingface.co/docs/transformers/index},  etc.   Then,  each  algorithm
requires necessary information, such as authors, definitions, reference information, and programming
libraries. The  algorithm structure of  basic information or  transformation is provided  in Section
\ref{sec:Ontology}. Moreover, well-known algorithms were defined in the AgriComO ontology. Similarly,
all evaluation metrics  for data mining processes  and common pre-trained models  were collected and
put into the ontology.  In the current  implementation, AgriComO contains 176 algorithms, 110 pre-trained
models, and 51 evaluation metrics, as shown in Table \ref{table:AgriComO:MainIndices}.

\setlength{\belowcaptionskip}{-15pt}
\vspace{-5mm}
\begin{table}[h]
  \centering
  \caption{Statistics of Main Indices and Transformations in AgriComO}{
    \label{table:AgriComO:MainIndices}%
    \begin{tabular}{llr}
      \hline\noalign{\smallskip}
      \textbf{Indices}  & \textbf{Examples} & \textbf{Count}\\
      \hline\noalign{\smallskip}
      Agronomic Indices      & Soil, weeds, nutrient, water indices       & 310 \\
      Agroclimatic Indices   & GDD, AFD, CDD, FX                          & 172 \\
      Vegetation Indices     & LAI, NDVI, GDVI, RVI, SAVI, SAVO           & 322 \\
      Ecological Indices     & Water Index: WZI, WFI                      &  15 \\
      Pretrain Models        & GoogLeNet, AlexNet                         & 110 \\
      Mining Algorithms      & Linear, DT, ANN, RF, LSTM, LASSO           & 176 \\
      Evaluation Metrics     & Accuracy, CCR, Jscore, F1, R2, RSME        &  51 \\
      Data Transformations   & RGB or BW for Colour, SoilPH Types         & 326 \\
      \hline\noalign{\smallskip}
      \textbf{Total}         &                                           & {\bf 1,310}\\
      \hline\noalign{\smallskip}
    \end{tabular}}%
\end{table}%
%\vspace{-10mm}

%-------------------------------------------------------------------------------
\section{Conclusion and Future Work}
\label{sec:conc}

We proposed  the OAK4XAI  model that  consists of an  ontology-based knowledge  map and  a knowledge
management system.  The OAK4XAI  architecture has the  potential to be  expanded to  other knowledge
domains.  The knowledge  items can  easily be  imported into  the knowledge  management system.  The
knowledge management system  implements a knowledge browser  to access and explore  knowledge of any
kind and with different levels of interpretations depending on the DM applications. 

We defined an agriculture ontology, AgriComO, and populated it with the most well-known algorithm for
mining agricultural data. AgriComO provides definitions and information for concepts, algorithms, and
states in crop farming.  We believe that the proposed model and its  ontology can provide consistent
information and explanation not  only for DM applications in agriculture but  also for other domains
by pre-defining their corresponding ontology. 

In future work, we will focus on two areas:  a) implement the explanation interface as a service, to
interact  with  several  user groups.  It  will  take  user  questions  as input  and  retrieve  the
corresponding knowledge,  and b)  extend the  proposed model  to include  several ML  algorithms for
prediction based on explainable approaches.

Moreover, we plan to  represent the result of several predictive algorithms  in agriculture, such as
decision    trees,   Bayesian    analysis,   or    support   vector    machines   with    rule-based
extraction.  Representations will  be compatible  with  different explanation  interfaces, and  then
develop a reasoning module to decide the most appropriate explanation. 

\paragraph{\bf Acknowledgment} 
This  work  is  part of  CONSUS  and  is  supported  by  the SFI  Strategic  Partnerships  Programme
(16/SPP/3296) and is co-funded by Origin Enterprises Plc. 

\bibliography{OAK4XAI}{}
\bibliographystyle{plain}

\end{document}